\documentclass{article}
\usepackage{ijcai22}

\usepackage{times}

\usepackage{soul}
\usepackage{url}
\usepackage[hidelinks]{hyperref}
\usepackage[utf8]{inputenc}
\usepackage[small]{caption}
\usepackage{subcaption}
\usepackage{graphicx}
\usepackage{amsmath}
\usepackage{booktabs}
\usepackage{color}

\definecolor{yun}{rgb}{0.858, 0.188, 0.478}

\definecolor{lei}{rgb}{0.2, 0.6, 1}

\title{CARBEN: Composite Adversarial Robustness Benchmark}

\author{
Lei Hsiung$^1$\footnote{Contact Author}\and
Yun-Yun Tsai$^2$\and
Pin-Yu Chen$^3$\And
Tsung-Yi Ho$^1$\\
\affiliations
$^1$National Tsing Hua University \\
$^2$Columbia University \\
$^3$IBM Research\\
\emails
hsiung@m109.nthu.edu.tw,
yt2781@columbia.edu,
pin-yu.chen@ibm.com,
tyho@cs.nthu.edu.tw
}

\begin{document}

\maketitle

\begin{abstract}
Prior literature on adversarial attack methods has mainly focused on attacking with and defending against a single threat model, e.g., perturbations bounded in Lp ball. However, multiple threat models can be combined into composite perturbations. One such approach, composite adversarial attack (CAA), not only expands the perturbable space of the image, but also may be overlooked by current modes of robustness evaluation. This paper demonstrates how CAA's attack order affects the resulting image, and provides real-time inferences of different models, which will facilitate users' configuration of the parameters of the attack level and their rapid evaluation of model prediction. A leaderboard to benchmark adversarial robustness against CAA is also introduced.

\end{abstract}

\section{Background}
Deep neural networks (DNNs) have transformed computer vision and have been used in many fields, including aviation, climate forecasting, and medicine, among others. However, when a DNN encounters carefully crafted images, it can exhibit various vulnerabilities: for instance, lack of robustness in the face of \textit{adversarial attack} \cite{szegedy2013intriguing}. By utilizing adversarial attacks to optimize perturbations, one can intentionally derive a perturbed sample from a normal image, and make it imperceptible to human beings.
In other words, despite the original and adversarial images looking very much alike to us, the latter can lead well-trained DNN models to make wrong predictions.

Previous studies have sought to create bounded perturbations in a metric manner \cite{goodfellow2015explaining,ChenAndSharma2018EAD}.  Most such work has focused on $\ell_{p}$-norm perturbation (i.e., $\ell_{1}$, $\ell_{2}$, and $\ell_{\infty}$) and utilized gradient-based optimization – i.e., fast gradient sign method (FGSM), projected gradient descent (PGD), or C\&W – to effectively generate the adversarial example. However, it is possible to extend adversarial perturbations beyond the $\ell_{p}$-norm bounds. For instance, Laidlaw et al. generated their perturbation in a perceptual distance metric \cite{laidlaw2021perceptual}, and Hosseini and Poovendran proposed a method of modifying images semantically to create semantic adversarial examples \cite{Hosseini_2018_CVPR_Workshops}. 
Additionally, Mao et al. studied adversarial examples generated from multiple-threat models, and demonstrated that they were more effective than single-threat ones against DNN targets \cite{mao2021composite}.

\begin{figure}[t]
    \centering
    \includegraphics[width=\linewidth, trim = 0.1cm 0.25cm 0.1cm 0.25cm, clip]{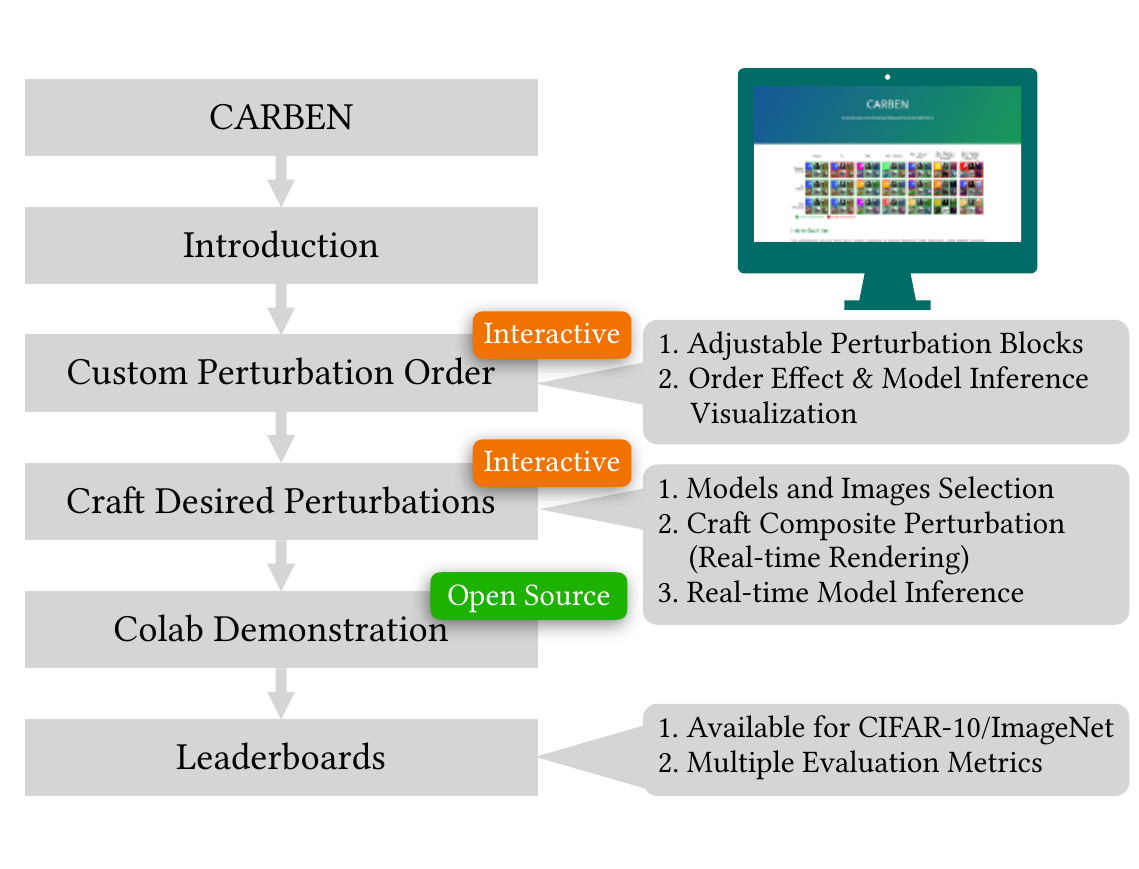}
    \caption{CARBEN overview. Browse on: \href{https://hsiung.cc/CARBEN/}{hsiung.cc/CARBEN}}
    \label{fig:demonstration-overview}
\end{figure}

Recently, Tsai et al. combined the $\ell_\infty$-norm and semantic perturbations (i.e., hue, saturation, rotation, brightness, and contrast), and proposed a novel approach -- composite adversarial attack (CAA) -- capable of generating unified adversarial examples \cite{tsai2022compositional}. The main differences between CAA and previously proposed perturbations are a) that CAA incorporates several threat models simultaneously, and b) that CAA's adversarial examples are semantically similar and/or natural-looking, but nevertheless result in large differences in $\ell_{p}$-norm measures.

Various defense strategies have also recently been proposed. For example, adversarial training (AT) is one of the most efficient ways to defend against adversarial attacks. However, recent results showed that $\ell_\infty$-robust models (e.g., \cite{madry2018towards}) might become fragile when they encounter composite perturbations \cite{tsai2022compositional}. Motivated by this limitation in $\ell_\infty$-centric robustness, generalized adversarial training (GAT) overcomes this weakness and shows the robustness against a variety of composite perturbations \cite{tsai2022compositional}.

To systematically track the progress of adversarial robustness, \cite{croce2021robustbench} created a leaderboard and benchmarks for 120+ state-of-the-art models’ performance on an image-classification task conducted under $\ell_{p}$-threat and common corruptions. 
However, their approach did not cover evaluation of robustness against semantic attacks or composite perturbations, and these absences could potentially have led to bias in their interpretation and ranking of DNN models.
To bridge this gap, familiarize other researchers with the concept of composite adversarial robustness, and ultimately, create more trustworthy AI, we developed a browser-based composite perturbation generation demo, CARBEN (composite adversarial robustness benchmark).

\begin{figure}[t]
    \centering
    \includegraphics[width=\linewidth]{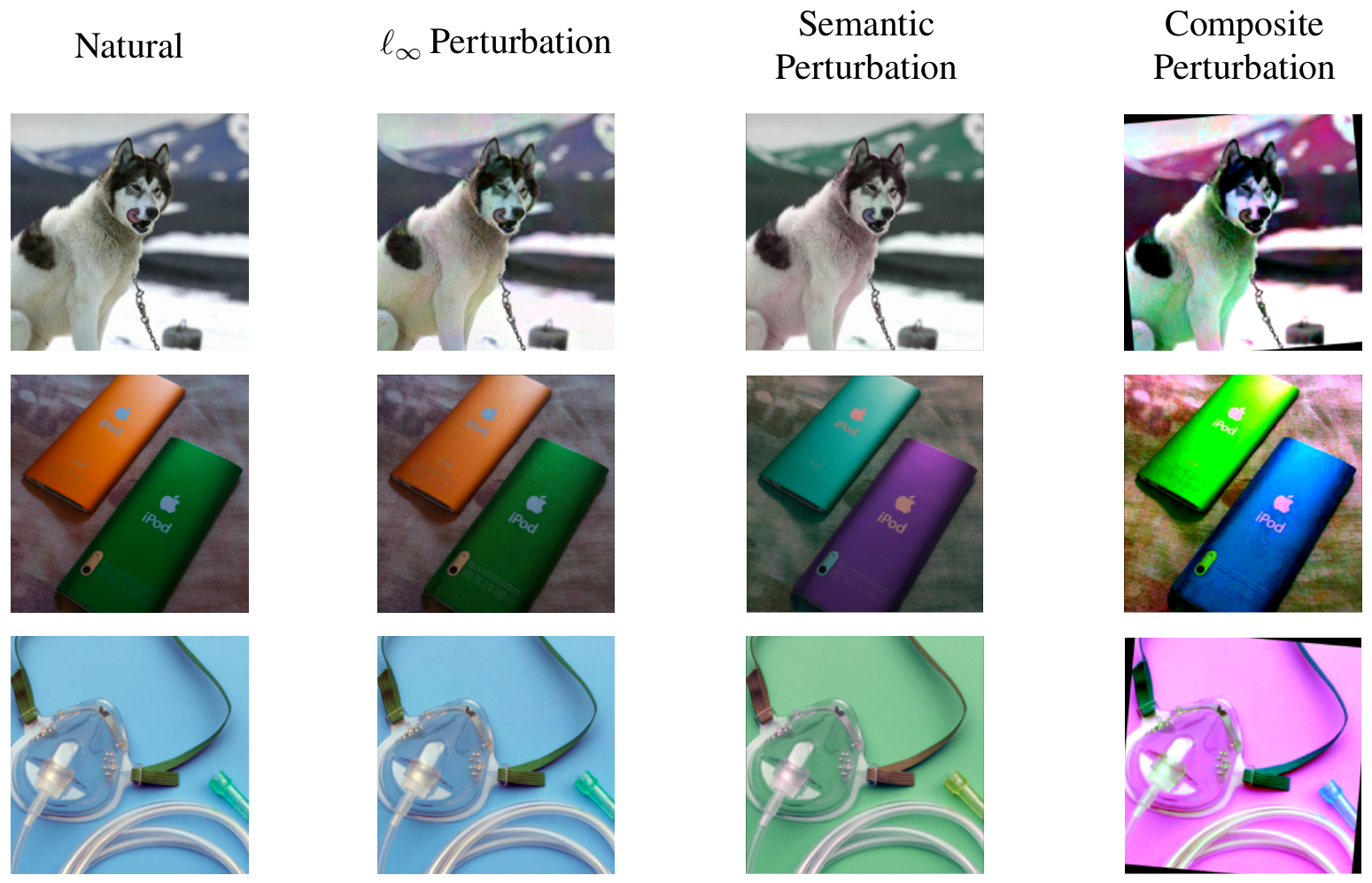}
    \caption{Examples of different perturbation types}
    \label{fig:perturbation-types}
\end{figure}

As shown in Figure \ref{fig:demonstration-overview}, CARBEN is a web application featuring \textit{interactive} and \textit{real-time} perturbation panels in which its users render the perturbed image and can see the models’ predictions in real time.
Figure \ref{fig:perturbation-types} presents examples of several perturbation types. CARBEN also includes an interactive section that lets users generate images in any attack combination. We have also created a leaderboard that tracks the accuracy of robustness against CAA.

\section{System Design}
The main purpose of CARBEN is to visualize the mechanism of generating composite perturbations, and thereby help its users to understand how models change their predictions or confidences when under adversarial attack. In this demonstration, users can adjust the attack parameters manually, to explore the effects such adjustments will have, and gain valuable hands-on experience of generating desired examples. Once users understand the concept of CAA and attack ordering, they can use our Google Colab demonstration and the automated CAA to generate a set of composite adversarial examples for robustness evaluation, and report their model-performance results on our leaderboard.


\subsection{Composite Perturbations in Custom Order}
To demonstrate how attack order can affect the outcomes of composite perturbations, we designed CARBEN to allow its users to explore order effects. Figure \ref{fig:custom-order} shows some perturbation examples in different orders of CAA. We pre-generated the images with all attack combinations and orders, and one can move the arrow-shaped attack blocks to the desired position to see the images in that selected order.
Furthermore, we also presented the highest confidence scores for each instance, measured with an $\ell_{\infty}$-robust model. This visualization also illustrates why previous models are more susceptible to semantic perturbations and lead to erroneous predictions when it comes to semantic and compositional scenarios.

\definecolor{correct}{rgb}{0, 0.615, 0}
\definecolor{incorrect}{rgb}{1, 0.333, 0.333}

\begin{figure}[t]
    \begin{subfigure}[b]{\linewidth}
        \centering
        \includegraphics[width=\textwidth]{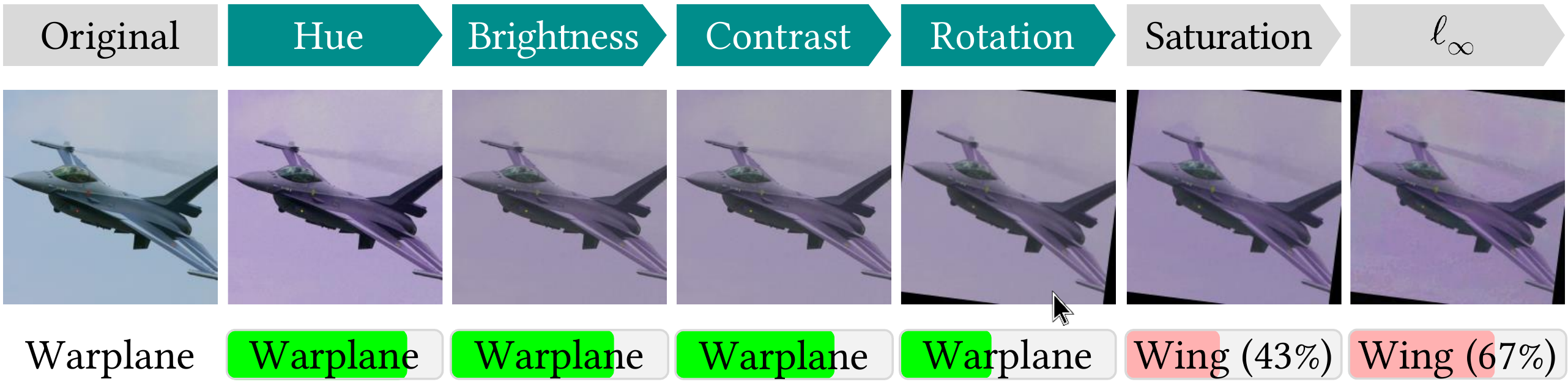}
        \caption{Perturbation block highlight. If the cursor hovers over the image below, the perturbation blocks are highlighted, indicating which perturbations are included in the image.}
        \label{fig:attack-order-example-warplane}
    \end{subfigure}
    \hfill
    \begin{subfigure}[b]{\linewidth}
        \centering
        \includegraphics[width=\textwidth]{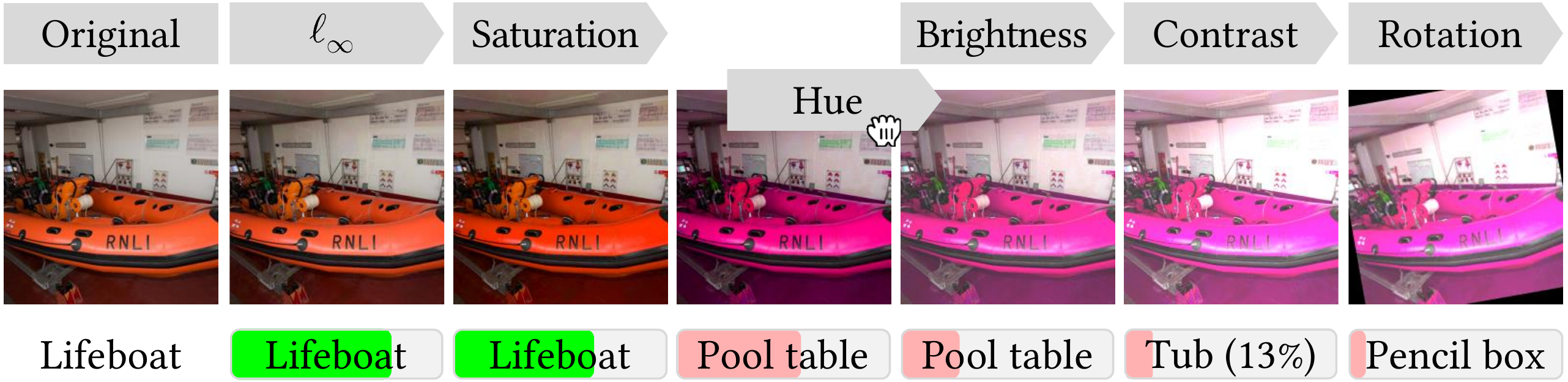}
        \caption{Adjustment of perturbation order. Users can change the perturbation order by exchanging perturbation blocks. }
        \label{fig:attack-order-example-lifeboat}
    \end{subfigure}
    \caption{Perturbed images, with each row representing a series of consecutive perturbations in a different specified attack order. Several samples and the inference results from an $\ell_\infty$-robust model are presented, and the confidence bar is marked in \textcolor{correct}{green} (\textcolor{incorrect}{red}) if the prediction is correct (incorrect).}
    \label{fig:custom-order}
\end{figure}


\begin{figure}[!t]
    \centering
    \includegraphics[width=\linewidth]{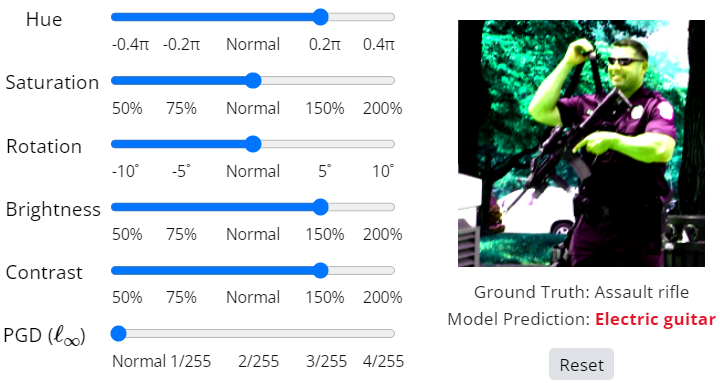}
    \caption{CARBEN’s interactive perturbation panel. Users can change the attack level and render the perturbed image on the canvas at right. Real-time model inference about the resulting instance is also provided, and marked in \textcolor{correct}{green} (\textcolor{incorrect}{red}) if it gives the correct (incorrect) prediction. The model used here is naturally trained model from \textit{torchvision} package.}
    \label{fig:perturbation-panel}
\end{figure}

\subsection{Perturbation Panel}
Because each attack component has its own perturbation level (or attack parameter), CARBEN also includes a perturbation panel, as shown in  Figure \ref{fig:perturbation-panel}, in which users can create composite perturbations on images. More specifically, a user can slide the bar to specify the attack level, and if it is set to \textit{Normal}, the attack will be disabled. However, it should be noted that for purposes of this CARBEN feature, the attack order is fixed as: $\ell_\infty\to\text{Hue}\to\text{Saturation}\to\text{Brightness}\to\text{Contrast}\to\text{Rotation}$. 

In addition, we provide several ImageNet examples and real-time model prediction on the user-generated image, including a standard training model, $\ell_\infty$-robust model, and GAT. Potentially, CARBEN could also be extended to support uploading images from user's phone or computer.

\subsection{Colab Demonstration and Leaderboard}
\paragraph{Colab Demonstration.} The opensource code for generating a composite adversarial example with an automatically optimized attack order is currently available on GitHub.\footnote{\url{https://github.com/twweeb/composite-adv}}
We provide a step-by-step tutorial on Colab to guide users on how to execute this CAA in the notebook and see the results.  The CIFAR-10 dataset was used for the experiment in our Colab notebook. Because CAA is implemented based on gradient optimization, computational cost and computing time will increase significantly as the number of enabled attacks increases. Therefore, we recommend using NVIDIA 2080Ti or above GPU to get better efficiency.

\paragraph{Benchmark and Leaderboard.}
When we compared the robustness rankings of the top 10 models on RobustBench (CIFAR-10, $\ell_{\infty}$) \cite{croce2021robustbench}, the results show that rankings between Auto-Attack and CAA (Full attacks) have a low correlation, suggesting that only considering perturbations in $\ell_p$ ball for robustness evaluation is biased and incomplete. Statistically, the Spearman’s rank correlation coefficients between Auto-Attack and CAA (Semantic attacks and Full attacks) are as follows: 0.16 for semantic attacks, and 0.38 for full attacks. 

\begin{figure}[t]
    \centering
    \begin{subfigure}[b]{\linewidth}
        \centering
        \includegraphics[width=\textwidth]{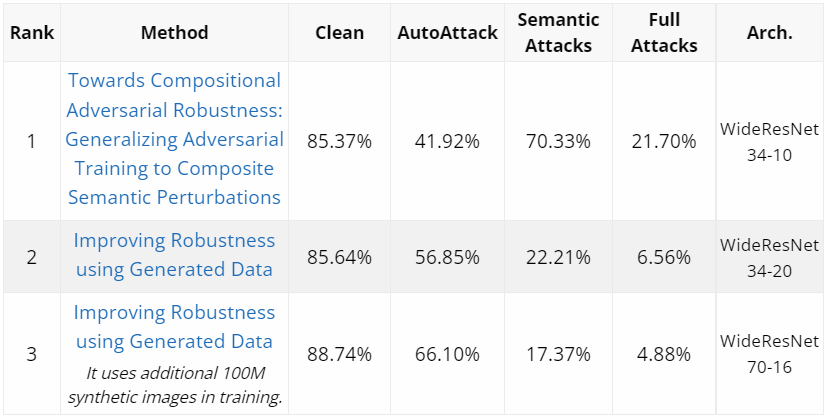}
        \caption{CIFAR-10}
        \label{fig:leaderboard-cifar10}
    \end{subfigure}
    \begin{subfigure}[b]{\linewidth}
        \centering
        \includegraphics[width=\textwidth]{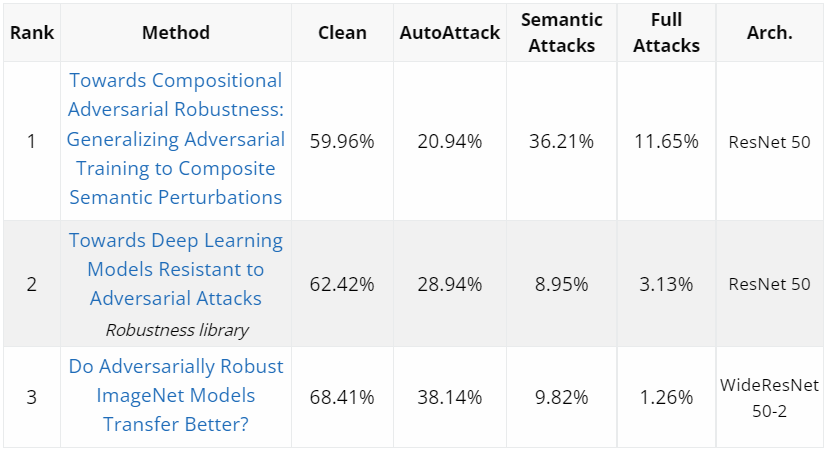}
        \caption{ImageNet}
        \label{fig:leaderboard-imagenet}
    \end{subfigure}
    \caption{The current top three entries on our leaderboard for model accuracy and robustness evaluation}
    \label{fig:leaderboard}
\end{figure}

To provide a complete robustness evaluation, we sought to offer several metrics for measuring the model performance. Therefore, we maintain leaderboards to facilitate tracking and benchmarking adversarial robustness progress across literature. Inspired from RobustBench, CARBEN's leaderboard focuses on tracking the robustness of model's robust accuracy not merely AutoAttack but also two CAAs (Semantic/Full attacks). Figure \ref{fig:leaderboard} shows the top three models on our leaderboard, as evaluated using CIFAR-10 and ImageNet datasets. In our leaderboard, models are ranked according to their \textit{Full Attacks robust accuracy}; their architectures and papers are also listed.

In this leaderboard, we focus on "white-box" scenarios in which the attacker has all knowledge of the models. We have provided similar entries to those in the RobustBench leaderboard, and hereby solicit model submissions to compete against composite perturbations in our leaderboard.

\section{Potential Impacts}
CARBEN is believed to be the first demo aimed at explaining composite perturbations, and demonstrates the effects on model prediction of altering the attack order. As such, it can help its users gain valuable insights into composite adversarial robustness, and thereby accelerate research devoted to the design of robust models.

Looking beyond conventional $\ell_p$-norm perturbations, our demo focuses on semantic adversarial attacks and composite perturbations, which are closer to real-life scenarios demanding robustness. For example, a brightness attack may happen during adjustments to a camera’s aperture, and a hue-and-saturation attack may occur when the camera lens is fitted with a filter or mask. Our CARBEN demo and leaderboard offer unprecedentedly realistic and comprehensive robustness assessment. Our web-based demonstration and hands-on interaction can also help to build awareness of AI and robustness, and thus serve as an education toolkit for trustworthiness in AI.


\section{Conclusion}

This demonstration enables CARBEN users to gain familiarity with CAAs. Its design features interactive sections, and provides Colab tutorials for both manual and automated composite-attack generation. Our robustness evaluation and leaderboard are believed to be the first attempts to benchmark model performance against complex and realistic threats beyond small-norm and single-type perturbations.


\section*{Acknowledgments}
This work is supported by Ministry of Science and Technology, Taiwan (MOST 111-2218-E-005-006-MBK).

\bibliographystyle{named}
\bibliography{ijcai22}

\end{document}